\documentclass[letterpaper, 10 pt, conference]{ieeeconf}
\IEEEoverridecommandlockouts
\pdfoutput=1
\let\chapter\section

\renewcommand\dbltextfloatsep{3pt}
\renewcommand\dblfloatsep{3pt}
\renewcommand\textfloatsep{3pt}
\usepackage{graphicx}
\usepackage{pifont}

\usepackage{booktabs}

\usepackage{subfig}
\usepackage[linesnumbered,ruled]{algorithm2e}
\usepackage[format=plain,labelsep=period,font=footnotesize]{caption}

\let\labelindent\relax
\usepackage{enumitem}
\usepackage[bbgreekl]{mathbbol}
\usepackage{amsfonts}

\DeclareSymbolFontAlphabet{\mathbb}{AMSb}
\DeclareSymbolFontAlphabet{\mathbbl}{bbold}

\usepackage{cite}
\usepackage{amsmath}
\usepackage{amssymb,mathrsfs}
\usepackage{array}

\usepackage{hyperref}
\usepackage{indentfirst}

\newtheorem{problem}{Problem}

\makeindex             

\usepackage[figuresright]{rotating}
\newcommand{\compresslist}{%
\setlength{\itemsep}{1pt}%
\setlength{\parskip}{0pt}%
\setlength{\parsep}{0pt}%
}

\begin{document}
\bstctlcite{IEEEexample:BSTcontrol}

\title{\LARGE \bf On the Use of the Observability Gramian\\ for Partially Observed Robotic Path Planning Problems$ ^{\tiny{*}} $}

\author
{
Mohammadhussein Rafieisakhaei$^{1}$, Suman Chakravorty$^{2}$ and P. R. Kumar$^{1}$
\thanks{*This material is based upon work partially supported by NSF under Contract Nos. CNS-1646449 and Science \& Technology Center Grant CCF-0939370, the U.S. Army Research Office under Contract No. W911NF-15-1-0279, and NPRP grant NPRP 8-1531-2-651 from the Qatar National Research Fund, a member of Qatar Foundation.}
\thanks{$^{1}$M. Rafieisakhaei and P. R. Kumar are with the Department of Electrical and Computer Engineering, and $^{2}$S. Chakravorty is with the Department of Aerospace Engineering, Texas A\&M University, College Station, Texas, 77840 USA.
        \{\tt\small mrafieis, schakrav, prk@tamu.edu\}}%
}

\maketitle
\thispagestyle{empty}
\pagestyle{empty}
\begin{abstract}
Optimizing measures of the observability Gramian as a surrogate for the estimation performance may provide irrelevant or misleading trajectories for planning under observation uncertainty.
\end{abstract}

\section{Introduction}\label{sec:Introduction}
The Observability Gramian (OG) is used to determine the observability of a deterministic linear time-varying system \cite{maybeck1982stochastic,yasuda1991assigning,vaidya2007observability}. For such systems, the properties of the OG have been well-studied \cite{maybeck1982stochastic, southall1998controllability, krener2009measures}. When sensors provide noisy stochastic measurements, the state is only partially observed. The general problem of planning under process and observation uncertainties has been formulated as such a stochastic control problem with noisy observations. The solution of this problem provides an optimal policy via the Hamilton-Jacobi-Bellman equation \cite{Kumar-book-86,Bertsekas07}. However, the computational hurdle for finding a solution to these equations has necessitated the study of a variety of methods to approximate the solution \cite{rafieisakhaei2016feedbackICRA, platt2013convex,Platt10ML,van2011lqg}. One approach has been to maximize the estimation performance by planning for trajectories that can exploit the properties of observation, process and a priori models. We examine the appropriateness or lack thereof of methods based on the OG, and show that they can provide misleading trajectories.

Borrowed from deterministic control theory, the OG has been exploited in order to provide \textit{more observable} trajectories, particularly in trajectory planing problems \cite{georges2011energy, hinson2014observability, hinson2013path, quenzer2014observability, travers2015use, devries2015wake, devries2013observability}. In the special case of a diagonal observation covariance with the same uncertainty level in each direction \cite{maybeck1982stochastic}, the Standard Fisher Information Matrix (SFIM) does reduce to the OG. Indeed the usage of the OG in filtering problems has been justified through its connections to the SFIM and its relations to the parameter estimation problem \cite{singh2005determining,hinson2014observability}. In fact, tailored to the parameter estimation problem, the SFIM only addresses the amount of information in the measurements alone \cite{maybeck1982stochastic}, and neglects both the prior information and process uncertainty. Closely-related approaches are the methods that base their planning on the observation model or the likelihood function \cite{rafieisakhaei2016feedbackICRA, platt2012non}, and the analysis of this paper can be helpful in providing a better understanding of those problems. 

In contrast, the \textit{Posterior} FIM (PFIM), whose inverse coincides with the Posterior Cram\'{e}r-Rao Lower Bound for the estimation uncertainty in a general stochastic problem \cite{tichavsky1998posterior}, can capture the history of evolution of uncertainty in the problem. In particular, for a linear system, it has been shown that the Riccati equations for the covariance evolution of the state estimation resulting from the Kalman Filter (KF) coincide with evolution of the PFIM in the form of the inverse covariance or the information filter \cite{tichavsky1998posterior, thacker1996tutorial,bastani2018fault, lei2010fisher}. Indeed, it is only this measure that can capture the entire information required to calculate the optimal policy along with the nominal trajectory of a stochastic system. It is therefore no surprise that these equations provide the evolution of the information state (the posterior or conditional distribution of the state given the entire history of actions and observations) as the sufficient statistic for decision-making through the Bayesian filtering equations.

In this paper, through a series of analytic and numerical examples, we show that the observability Gramian does not generally provide an appropriate solution for the problem of planning under uncertain observations. We provide examples for two commonly used nonlinear observation models including the range and squared-range observation models that provide noisy information regarding the state of the system with respect to a set of information sources or landmarks. The examples show that the OG is insensitive to the uncertainty parameters of the problem, with none of the three main covariances, i.e., process, observation or initial, appearing quantitatively. Similarly, we show that the SFIM also suffers the same problems as the OG.

The numerical examples illustrate the performance of simple planning problems when a measure of the OG (or SFIM in special case) is utilized as the optimization objective. In these examples, the trace of the error covariance, which represents the sum of mean squared errors along the trajectory, is used as the measure of performance of trajectory. In each example, the OG-based trajectory's performance is evaluated against both an initial trivial path and the optimized path with respect to the trace of the covariance. The results indicate that for all three models there are situations where the OG-based trajectory can perform significantly poorly with respect to these two trajectories, including even the initial trivial path. In some situations the trajectories produced are qualitatively similar, while their estimation performances are very different.

On the other hand, due to some very special circumstances OG-based planning may sometimes be close to the optimal outcome, and we provide such an example too. The above examples shows that OG-based planning is not reliable. One of the main reasons for usage of the OG-based method has been its relatively simpler computation, in comparison to the Riccati equation. However, we show that while there is a constant-factor computational difference in terms of the matrix calculations, a careful formulation of the original problem can lead to the same ``order" of computation as the OG-based problem. 

We introduce the preliminary notations and definitions of the Gramian and some OG-based measures in the next section. Then, we proceed to the analytic examples in Section \ref{sec:Examples with Analytic Calculations}. In Section \ref{sec:Planning Problems}, we provide several formulations of planning problems and describe the numerical simulation results. 

\section{Preliminaries}\label{sec:General Problem}
We begin with some preliminary definitions.

\emph{Process and observation models:} Let $\mathbf{x}\in \mathbb{X} \subset\mathbb{R}^{n_x}$, $\mathbf{u}\in \mathbb{U}\subset\mathbb{R}^{n_u}$ and $\mathbf{z}\in \mathbb{Z}\subset\mathbb{R}^{n_z}$ denote the state, control and observation vectors, respectively. We use boldface variables to denote the vectors in lower case and matrices in upper case, respectively. Let $\mathbf{f}:\mathbb{X}\times\mathbb{U}\times\mathbb{R}^{n_u}\rightarrow\mathbb{X}$ and $\mathbf{h}:\mathbb{X}\rightarrow\mathbb{Z}$ denote the general process and observation models:  
\begin{subequations}\label{eq:non-linear system equations}
\begin{alignat}{2}
\mathbf{x}_{t+1}&=\mathbf{f}(\mathbf{x}_{t},\mathbf{u}_{t},\boldsymbol{\omega}_{t}), ~~&&\boldsymbol{\omega}_{t}\sim \mathcal{N}(\mathbf{0}, \boldsymbol{\Sigma}_{\boldsymbol{\omega}}),\label{eq:linear-sys-general-app-2}\\
\mathbf{z}_{t}&=\mathbf{h}(\mathbf{x}_{t},\boldsymbol{\nu}_{t}),~~ && \boldsymbol{\nu}_{t}\sim \mathcal{N}(\mathbf{0}, \boldsymbol{\Sigma}_{\boldsymbol{\nu}}),
\end{alignat}
\end{subequations}
where $ \{\boldsymbol{\omega}_t\} $ and $ \{\boldsymbol{\nu}_t\} $ are zero mean independent, identically distributed (i.i.d.) mutually independent random sequences, with $ \mathcal{N}(\mathbf{m}, \boldsymbol{\Sigma}) $ denoting a normal distribution with mean $ \mathbf{m} $ and covariance $ \boldsymbol{\Sigma} $.

\textit{Parameterized Trajectories}: Starting with an initial estimate, $ \mathbf{x}^{p}_{0} := \hat{\mathbf{x}}_{0} $, and using a set of unknown control inputs $ \{ \mathbf{u}^{p}_{t}\}_{t=0}^{K-1} $, we parametrize the possible feasible nominal trajectories of the system:
\begin{align*}
\mathbf{x}^{p}_{t+1}&:=\mathbf{f}(\mathbf{x}^{p}_t, \mathbf{u}^{p}_t, \mathbf{0}),~~ 0\le t \le K\!-\!1,\\
\mathbf{z}^{p}_{t}&:=\mathbf{h}(\mathbf{x}^{p}_{t}, \mathbf{0}),~~ 1\le t \le K.
\end{align*}

\textit{Linearization of the system equations:} We linearize the nonlinear motion and observation models of equation \eqref{eq:non-linear system equations} about the parametrized trajectory:
\begin{subequations}\label{eq:linearized system}
\begin{align}
\tilde{\mathbf{x}}_{t+1}&=\mathbf{A}_t\tilde{\mathbf{x}}_t + \mathbf{B}_t\tilde{\mathbf{u}}_t +\mathbf{G}_t\boldsymbol{\omega}_t,\\
\tilde{\mathbf{z}}_{t}&=\mathbf{H}_t\tilde{\mathbf{x}}_t+\mathbf{M}_t\boldsymbol{\nu}_t,
\end{align}
\end{subequations}
where $ \tilde{\mathbf{x}}_{t}\!:=\!\mathbf{x}_t\!-\!\mathbf{x}^{p}_{t} $, $ \tilde{\mathbf{u}}_{t}\!:=\!\mathbf{u}_t\!-\!\mathbf{u}^{p}_{t} $, and $ \tilde{\mathbf{z}}_{t}\!:=\!\mathbf{z}_t\!-\!\mathbf{z}^{p}_{t} $ denote the state, control and observation errors, respectively, and
\begin{align*}
&\mathbf{A}_t:=\nabla_{\mathbf{x}} \mathbf{f}(\mathbf{x},\mathbf{u},\boldsymbol{\omega})|_{ \mathbf{x}^{p}_{t}, \mathbf{u}^{p}_{t}, \mathbf{0} }, \mathbf{B}_t:=\nabla_{\mathbf{u}} \mathbf{f}(\mathbf{x},\mathbf{u},\boldsymbol{\omega})|_{ \mathbf{x}^{p}_{t}, \mathbf{u}^{p}_{t}, \mathbf{0} },
\\&\mathbf{G}_t:=\nabla_{\boldsymbol{\omega}} \mathbf{f}(\mathbf{x},\mathbf{u},\boldsymbol{\omega})|_{ \mathbf{x}^{p}_{t}, \mathbf{u}^{p}_{t}, \mathbf{0} },
\mathbf{H}_t(\mathbf{x}^{p}_{t}):=\nabla_{\mathbf{x}} \mathbf{h}(\mathbf{x},\boldsymbol{\nu})|_{ \mathbf{x}^{p}_{t},\mathbf{0} },  \\&\mathbf{M}_t(\mathbf{x}^{p}_{t}):=\nabla_{\boldsymbol{\nu}} \mathbf{h}(\mathbf{x},\boldsymbol{\nu})|_{ \mathbf{x}^{p}_{t},\mathbf{0} }.
\end{align*}
Note that $ \{\mathbf{x}^{p}_{t}\}_{t=0}^{K} $, $ \{\mathbf{z}^{p}_{t}\}_{t=0}^{K} $, and the Jacobian matrices change upon change of the underlying control inputs $ \{ \mathbf{u}^{p}_{t}\}_{t=0}^{K-1} $.

\subsection{Observability Gramian}
\textit{Observability Gramian:} Let $ \tilde{\mathbf{A}}_{t}:=\Pi_{\tau=0}^{t}\mathbf{A}_{\tau} $ denote the transition matrix of the linearized system of \eqref{eq:linearized system} starting from time $ 0 $. Then, the $ (K\!+\!1) $-step observability Gramian corresponding to the nominal trajectory is defined as:
\begin{align}\label{eq:OG}
\mathbf{Q}^{p}_{K+1}:=\sum_{t=0}^{K}\tilde{\mathbf{A}}^{T}_{t}\mathbf{H}^{T}_{t}\mathbf{H}_{t}\tilde{\mathbf{A}}_{t}.
\end{align}
The noise-less system of exactly linear equations is observable if and only if $ \mathrm{rank}(\mathbf{Q}^{p}_{n_x\!-\!1})=n_x $ \cite{maybeck1982stochastic}.

Note that as the control inputs $ \mathbf{u}^{p}_{t} $ change, $ \mathbf{Q}^{p}_{K+1} $ changes, as well. This has led to a variety of approaches to utilize the OG or some function of the OG as a measure to optimize in the trajectory optimization problems. One motivating factor, as mentioned above, is the low computational burden of computing the OG. Another motivating factor for using the OG is its proven role in determining the \textit{initial state}, $ \mathbf{x}^{p}_{0} $, i.e., observability property of a deterministic system. 
However, in the stochastic case, \textit{given} (partial) information around the \textit{initial} state, the goal is to find trajectories where the state becomes more observable along the trajectory (including, in particular, the final state, which may be important to goal-oriented problems, as opposed to the initial state). 

\textit{Measures of the Gramian:} In several papers, e.g., \cite{singh2005determining,hinson2014observability}, the following scalar measures of the OG have been used with various interpretations related to the uncertainty in the systems:
\begin{itemize}
\item Determinant of the inverse OG, $ \det((\mathbf{Q}^{p}_{K+1})^{-1})=\det^{-1}(\mathbf{Q}^{p}_{K+1}) $ (and sometimes logarithm of it);
\item Trace of the inverse OG, $ \mathrm{tr}((\mathbf{Q}^{p}_{K+1})^{-1}) $;
\item Negative trace of the OG, $ -\mathrm{tr}(\mathbf{Q}^{p}_{K+1}) $;
\item Inverse of the OG's minimum eigenvalue, $ \lambda_{\min}^{-1}(\mathbf{Q}^{p}_{K+1}) $;
\item Inverse of the OG's maximum eigenvalue, $ \lambda_{\max}^{-1}(\!\mathbf{Q}^{p}_{K+1}\!) $;
\item The condition number of the OG, $ \kappa(\mathbf{Q}^{p}_{K+1}) $.
\end{itemize}

\subsection{Standard Fisher Information Matrix}
A metric closely related to the Gramian is the SFIM the inverse of which is a lower bound on the minimum attainable estimation covariance for a parameter estimation problem as given by the Cram\'{e}r-Rao lower bound \cite{casella2002statistical}. The SFIM, $ \mathbf{F}_K $, for the system of equations \eqref{eq:linearized system} is calculated as \cite{maybeck1982stochastic}:
\begin{align}\label{eq:SFIM}
\mathbf{F}_K = \sum_{t=0}^{K}\tilde{\mathbf{A}}_{t}^{T}\mathbf{H}_{t}^{T}\boldsymbol{\Sigma}_{\boldsymbol{\nu}}^{-1}\mathbf{H}_{t}\tilde{\mathbf{A}}_{K}.
\end{align}
Note that in the special case $ \boldsymbol{\Sigma}_{\boldsymbol{\nu}}=\sigma\mathbf{I}_{n_z} $ with $ \sigma>0 $, the SFIM reduces to a weighted OG:
\begin{align}\label{eq:SFIM simplified}
\mathbf{F}_K =\frac{1}{\sigma} \sum_{t=0}^{K}\tilde{\mathbf{A}}_{t}^{T}\mathbf{H}_{t}^{T}\mathbf{H}_{t}\tilde{\mathbf{A}}_{K}=\frac{1}{\sigma}\mathbf{Q}^{p}_{K+1}.
\end{align}
\subsection{Covariance Evolution}
\textit{Information state:} The posterior distribution of $ \mathbf{x}_t $ given the history of actions and observations up to time-step $ t $, $ p_{\mathbf{X}_t|\mathbf{Z}_{0:t};\mathbf{U}_{0:t-1}, \mathbf{X}_0}(\mathbf{x}|\mathbf{z}_{0:t};\mathbf{u}_{0:t-1},\mathbf{x}_{0}) $, is referred to as the information state. It is a sufficient statistic for the stochastic control problem \cite{Kumar-book-86,Bertsekas07}. In the linear Gaussian case, the covariance evolution of the information state is specified by the Kalman filtering equations. The covariance evolution of the KF becomes deterministic once the underlying nominal linearization trajectory of the system equations is fixed:
\begin{subequations}\label{eq:riccati}
\begin{align}
\mathbf{P}^{-}_{t}&=\mathbf{A}_{t-1}\mathbf{P}^{+}_{t-1}\mathbf{A}_{t-1}^T+\mathbf{G}_{t-1}\boldsymbol{\Sigma}_{\boldsymbol{\omega}}\mathbf{G}_{t-1}^T\label{eq:riccati 1 },
\\\mathbf{S}_{t}&=\mathbf{H}_t\mathbf{P}^{-}_{t}\mathbf{H}_t^T+\mathbf{M}_t\boldsymbol{\Sigma}_{\boldsymbol{\nu}}\mathbf{M}_t^T,\label{eq:riccati 2 }
\\\mathbf{P}^{+}_{t}&=(\mathbf{I}-\mathbf{P}^{-}_{t}\mathbf{H}_t^T\mathbf{S}_{t}^{-1}\mathbf{H}_t)\mathbf{P}^{-}_{t}\label{eq:riccati 4 },~\mathbf{P}^{+}_{0}=\boldsymbol{\Sigma}_{\mathbf{x}_{0}}.
\end{align}
\end{subequations}

\section{Analytic Evaluation of OG-Based Designs}\label{sec:Examples with Analytic Calculations}
In this section, we provide two examples based on commonly used range and range-squared observation models in order to compare the amount of information and the different aspects of the models, such as stochasticity captured by the OG, the SFIM, and the PFIM equations.

\textit{System equations:} In the examples of this section, we have $\mathbf{x}\in\mathbb{R}^{2}$, $\mathbf{u}\in\mathbb{R}^{2}$, $\mathbf{z}\in\mathbb{R}$, and $ K>1 $. Moreover, the process and observation models are:
\begin{subequations}\label{eq:example system equations}
\begin{alignat}{2}
\mathbf{x}_{t+1}&=\mathbf{x}_{t}+\mathbf{u}_{t}+\boldsymbol{\omega}_{t}, ~~&&\boldsymbol{\omega}_{t}\sim \mathcal{N}(\mathbf{0}, \boldsymbol{\Sigma}_{\boldsymbol{\omega}}),\label{eq:example system equations-1}\\
{z}_{t}&=h(\mathbf{x}_{t})+{\nu}_{t},~~ && {\nu}_{t}\sim \mathcal{N}({0}, {\Sigma}_{{\nu}}),
\end{alignat}
\end{subequations}
where $ \{\boldsymbol{\omega}_t\} $ and $ \{{\nu}_t\} $ are zero mean i.i.d. random sequences that are mutually independent of each other, $ \mathbf{x}_{t}=[x_{t}, y_{t}]^{T} $, $ \boldsymbol{\Sigma}_{\boldsymbol{\omega}}=\mathrm{diag}(\sigma_{\boldsymbol{\omega}_{x}}, \sigma_{\boldsymbol{\omega}_{y}}) $, $ \Sigma_{\nu}=\sigma_{\nu} $, and the initial state is distributed as $ \mathbf{x}_{0}\sim \mathcal{N}(\hat{\mathbf{x}}_{0}, \boldsymbol{\Sigma}_{\mathbf{x}_{0}}) $, where $ \boldsymbol{\Sigma}_{\mathbf{x}_{0}}=\mathrm{diag}(\sigma_{x_0}, \sigma_{y_0}) $. Later in the simulations, we will consider a non-diagonal initial covariance, as well. Note that except for $ \mathbf{H}_{t} $, the other Jacobians of the above system are common to all examples, and are $ \mathbf{A}_{t}=\mathbf{I}_{2}, \mathbf{B}_{t}=\mathbf{I}_{2},
\mathbf{G}_{t}=\mathbf{I}_{2}$, and $
\mathbf{M}_{t}=\mathbf{I}_{1} $. As a result, $ \tilde{\mathbf{A}}_{t}=\mathbf{I}_{2}, t\ge 0 $.
\subsection{Range-Only Example}
Our first example involves an observation that acquires the range information relative to an information source located at the origin; i.e., $ h(\mathbf{x}_{t})=r_t=:\sqrt{(x_{t})^{2}+(y_{t})^{2}} $. The Jacobian of the observation model is $ \mathbf{H}_{t}=(\frac{x_{t}}{r_t}, \frac{y_{t}}{r_t}) $.

\textit{The OG calculations:} The OG for this system model is
\begin{align*} 
\mathbf{Q}^{p}_{K+1}&=\sum_{t=0}^{K}\begin{pmatrix}
    \frac{x^{2}_{t}}{r^2_t} & \frac{x_{t}y_{t}}{r^2_t} \\
    \frac{x_{t}y_{t}}{r^2_t} & \frac{y^{2}_{t}}{r^2_t}
  \end{pmatrix}.
\end{align*}
Note that the determinant of the OG is
\begin{align}\label{eq:det OG example 1}
\det(\mathbf{Q}^{p}_{K+1})=(\sum_{t=0}^{K}\frac{x^{2}_{t}}{r^2_t})(\sum_{t=0}^{K}\frac{y^{2}_{t}}{r^2_t})-(\sum_{t=0}^{K}\frac{x_{t}y_{t}}{r^2_t})^{2}>0,
\end{align}
which is positive using the Cauchy-Schwarz inequality, excluding situations where the trajectories of the two coordinates are linearly dependent (which includes a situation in which either coordinate's trajectory is entirely zero, or a situation that the state trajectory is a straight line whose extension can pass the origin). Therefore, except for these degenerate situations this system is observable. The trace of the OG is
\begin{align}\label{eq:trace OG rs}
\mathrm{tr}(\mathbf{Q}^{p}_{K+1})&=K+1,
\end{align}
which is a constant, insensitive to the underlying trajectory.

\textit{SFIM calculations:} Since the covariance of the observations is a constant and diagonal, the SFIM reduces to the form represented in equation \eqref{eq:SFIM simplified}, and $ \mathrm{tr}(\mathbf{F}_{K})=\sigma^{-1}_{\nu}\mathrm{tr}(\mathbf{Q}^{p}_{K+1})=\sigma^{-1}_{\nu}(K+1) $, which is a constant, insensitive to the underlying trajectory, just like the trace of the OG. In fact, the SFIM is a constant multiplier of the OG in all subsequent examples, as well.

\textit{Covariance of the estimation calculations:} The Riccati equations of \eqref{eq:riccati} for the evolution of the estimation covariance, in contrast, provide a different perspective than the OG and the SFIM. Starting from the initial covariance $ \mathbf{P}^{+}_{0}=\boldsymbol{\Sigma}_{\mathbf{x}_{0}} $,
the covariance ceases to be a diagonal after just one time step, and its trace $ t=1 $ is:
\begin{align}
\mathrm{tr}(\mathbf{P}^{+}_{1})=&\frac{(\sigma^{x}_{0}+\sigma^{x}_{\boldsymbol{\omega}})(\sigma^{y}_{0}+\sigma^{y}_{\boldsymbol{\omega}})+  (\sigma^{x}_{0}+\sigma^{x}_{\boldsymbol{\omega}}+\sigma^{y}_{0}+\sigma^{y}_{\boldsymbol{\omega}})\sigma_{\nu}} {(\sigma^{x}_{0}+\sigma^{x}_{\boldsymbol{\omega}})\frac{x^{2}_{t}}{r^2_t}+ (\sigma^{y}_{0}+\sigma^{y}_{\boldsymbol{\omega}})\frac{y^{2}_{t}}{r^2_t}
+\sigma_{\nu}}.\label{eq:trace cov}
\end{align}
Unlike in the case of the OG and the SFIM, minimization based on the covariance information is indeed sensitive to the underlying trajectory. In fact, this dependence is revealed after just one step of the Riccati equation's update.

\subsection{Range-Squared-Only Example}
Next, we consider a model that is often used in place of the range-only model and show that the behavior of the OG changes even by a simple squaring of the observation model. We have $ h(\mathbf{x}_{t})=\frac{1}{2}r^{2} $, with Jacobian given by $ \mathbf{H}_{t}=(x_{t}, y_{t}) $. 

\textit{The OG calculations:} The OG is
\begin{align*}
\mathbf{Q}^{p}_{K+1}&=\sum_{t=0}^{K}\begin{pmatrix}
    x^{2}_{t} & x_{t}y_{t} \\
    x_{t}y_{t} & y^{2}_{t}
  \end{pmatrix}.
\end{align*}
Its determinant is
\begin{align}
\det(\mathbf{Q}^{p}_{K+1})=(\sum_{t=0}^{K}x^{2}_{t})(\sum_{t=0}^{K}y^{2}_{t})-(\sum_{t=0}^{K}x_{t}y_{t})^{2}>0,
\end{align}
which is again taken to positive, assuming non-degenerateness. The trace of the OG is $ \mathrm{tr}(\mathbf{Q}^{p}_{K+1})=\sum_{t=0}^{K}r^2_t $,
maximizing which suggests trajectories that are \textit{farther} from the origin. We note that a simple squaring of the range produces exactly the opposite result, showing the inappropriateness of an OG-based design and requirement of a careful investigation with the covariance-based design. The SFIM measure also produces similar results.

\textit{Estimation covariance:} Similarly, given $ \mathbf{P}^{+}_{0}=\boldsymbol{\Sigma}_{\mathbf{x}_{0}} $,
the trace of the updated covariance at $ t=1 $ is:
\begin{align}
\mathrm{tr}(\mathbf{P}^{+}_{1})\!=\!&\frac{(\sigma^{x}_{0}+\sigma^{x}_{\boldsymbol{\omega}})(\sigma^{y}_{0}+\sigma^{y}_{\boldsymbol{\omega}})r^2_t \!+\! (\sigma^{x}_{0}+\sigma^{x}_{\boldsymbol{\omega}}+\sigma^{y}_{0}+\sigma^{y}_{\boldsymbol{\omega}})\sigma_{\nu}} {(\sigma^{x}_{0}+\sigma^{x}_{\boldsymbol{\omega}})x^{2}_{t}+ (\sigma^{y}_{0}+\sigma^{y}_{\boldsymbol{\omega}})y^{2}_{t}
+\sigma_{\nu}}\label{eq:trace cov example 3}.
\end{align}
This result also shows that, even after just one time step, the filtering equation provides very different and reasonable solutions than the OG or SFIM measures. Unlike the trace of the OG, this result does not suggest a uniform radial movement away from the origin; rather, it suggests paths that are dependent and sensitive to the direction of movement taking into account the uncertainty reductions in those directions.

\subsection{Observations}
Equations \eqref{eq:trace cov} and \eqref{eq:trace cov example 3}, which represent the trace of the PFIM in each case, provide far more valuable information than the any measure of the OG:
\begin{itemize}\compresslist
\item The trace of the updated PFIM depends on the underlying trajectory. In contrast, the trace of OG can become a constant regardless of the noise covariances, e.g., \eqref{eq:trace OG rs};
\item PFIM, takes into account the uncertainties in each direction. In contrast, the OG-based design can be insensitive to the directions involved;
\item The trace of the updated covariance is dependent on the previous covariance of the state estimation; 
\item The trace of covariance depends on both the observation and process noise covariances; and
\item PFIM's dependence on the process, observation and previous (history of uncertainty and prior) covariances is not uniform in each direction. However, measures of the OG are insensitive to such covariances.
\end{itemize} 

\section{Comparison of Trajectory Planning Approaches}\label{sec:Planning Problems}
In this section, we consider an optimal control problem that is common in path planning and control problems, particularly in robotic systems. We introduce the general problem and describe a commonly used surrogate open-loop optimal control problem whose cost function is a measure of the OG. Finally, we compare the above approaches with a trajectory optimization problem extending our previous work on the Trajectory-optimized Linear Quadratic Gaussian (T-LQG) in \cite{rafi2017ICRA,rafieisakhaei2016belief}, which optimizes the underlying trajectory of an LQG system aiming for the best estimation performance. This problem utilizes the trace of the covariance as the optimization objective and is accompanied by a separate feedback design implemented in the execution of the policy. In a companion paper, we prove the near-optimality of this framework under a small-noise assumption \cite{rafieisakhaei2016belief,rafieisakhaei2017SeparationCDC}.

\begin{problem}\label{problem:Stochastic Control Problem partially observed} \textup{\textbf{General Stochastic Control Problem}} Given $ \mathbf{x}_{0}\sim p(\mathbf{x}_{0}) $, solve for the optimal policy:
\begin{subequations}
\begin{align}\label{problem eq:Stochastic Control Problem partially observed}
\nonumber \min_{\pi}~\mathbb{E}[\sum_{t=0}^{K-1}&c_t^{\pi}(\mathbf{x}_t,\mathbf{u}_t)+c_K^{\pi}(\mathbf{x}_K)]
\\ s.t.~~\mathbf{x}_{t+1}&=\mathbf{f}(\mathbf{x}_{t},\mathbf{u}_t,\boldsymbol{\omega}_{t})
\\\mathbf{z}_{t}&=\mathbf{h}(\mathbf{x}_{t},\boldsymbol{\nu}_{t}),
\end{align}
\end{subequations}
where the optimization is over feasible policies, $ \mathbbl{\Pi} $, and:
\begin{itemize}\compresslist
\item $ \pi\in\mathbbl{\Pi} $, $ \pi:=\{\pi_{0}, \cdots, \pi_{t}\} $, $ \pi_{t} :\mathbb{Z}^{t+1}\rightarrow \mathbb{U} $ ;
\item $ \mathbf{u}_{t}=\pi_{t}(\mathbf{z}_{0:t}) $ specifies an action given the entire output of the system from the beginning up to time-step $ t $, $ \mathbf{z}_{0:t} $;
\item $ c^{\pi}_t(\cdot,\cdot):\mathbb{X}\times\mathbb{U}\rightarrow\mathbb{R} $ is the one-step cost function;
\item $ c_K^{\pi}(\cdot):\mathbb{X}\rightarrow\mathbb{R} $  denotes the terminal cost; and $ K>0 $.
\end{itemize}
\end{problem}

\begin{problem}\label{problem:Planning Problem OG}\textup{\textbf{OG-Based Trajectory Optimization Problem}} Solve for the optimal trajectory:
\begin{subequations}
\begin{align}
\nonumber \min_{\mathbf{u}^{p}_{0:K-1}} g(\mathbf{Q}^{p}_{K+1}&)+\sum\limits_{t=1}^{K}({\mathbf{u}}^{p}_{t-1})^{T}\mathbf{W}^{u}_{t}{\mathbf{u}}^{p}_{t-1}
\\s.t.~~\mathbf{x}^{p}_{t+1}&=\mathbf{f}(\mathbf{x}^{p}_t, \mathbf{u}^{p}_t, 0), ~0\!\le\! t\! \le\! K\!-\!1\label{eq:state propagation }
\\\mathbf{x}^{p}_{0} &= \mathbb{E}_{\mathbf{x}}[p(\mathbf{x}_{0})]\label{eq:initial mean }
\\|\!|\mathbf{x}^{p}_{K}&-\mathbf{x}_g|\!|_{2}<r_g\label{eq:terminal constraint }
\\|\!|\mathbf{u}^{p}_{t}|\!|_{2}&\le r_u, ~1\!\le\! t\! \le\! K,\label{eq:control contraint }
\end{align}
\end{subequations}
where the optimization is over feasible controls, $ g:\mathbb{R}^{n_x\times n_x}\rightarrow\mathbf{R} $ represents a specific operation on the OG, such as trace, determinant, etc., $ \mathbf{W}^{u}_{t}\succeq 0 $, $ r_u>0 $, and $ r_g>0 $ and $ \mathbf{x}_g\in\mathbb{X} $ specify the goal region.
\end{problem}

\begin{problem}\label{problem:Planning Problem}\textup{\textbf{T-LQG Planning Problem \cite{rafieisakhaei2016belief}}} Solve for the optimal linearization trajectory of the LQG policy:
\begin{subequations}
\begin{align}
\nonumber \min_{\mathbf{u}^{p}_{0:K-1}}\sum\limits_{t=1}^{K}[&\mathrm{tr}( \mathbf{P}^{+}_{\mathbf{b}^{p}_t})+({\mathbf{u}}^{p}_{t-1})^{T}\mathbf{W}^{u}_{t}{\mathbf{u}}^{p}_{t-1}]
\\s.t.~~\mathbf{P}^{-}_{t}&=\mathbf{A}_{t-1}\mathbf{P}^{+}_{t-1}\mathbf{A}_{t-1}^T+\mathbf{G}_{t-1}\boldsymbol{\Sigma}_{\boldsymbol{\omega}_{t-1}}\mathbf{G}_{t-1}^T\label{eq:riccati 1}
\\\mathbf{S}_{t}&=\mathbf{H}_t\mathbf{P}^{-}_{t}\mathbf{H}_t^T+\mathbf{M}_t\boldsymbol{\Sigma}_{\boldsymbol{\nu}_{t}}\mathbf{M}_t^T\label{eq:riccati 2}
\\\mathbf{P}^{+}_{t}&=(\mathbf{I}-\mathbf{P}^{-}_{t}\mathbf{H}_t^T\mathbf{S}_{t}^{-1}\mathbf{H}_t)\mathbf{P}^{-}_{t}\label{eq:riccati 4},~\mathbf{P}^{+}_{0}=\boldsymbol{\Sigma}_{\mathbf{x}_{0}} 
\\\mathbf{x}^{p}_{0} &= \mathbb{E}_{\mathbf{x}}[p(\mathbf{x}_{0})]\label{eq:initial mean}
\\\mathbf{x}^{p}_{t+1}&=\mathbf{f}(\mathbf{x}^{p}_t, \mathbf{u}^{p}_t, 0), ~0\!\le\! t\! \le\! K\!-\!1\label{eq:state propagation}
\\|\!|\mathbf{x}^{p}_{K}&-\mathbf{x}_g|\!|_{2}<r_g\label{eq:terminal constraint}
\\|\!|\mathbf{u}^{p}_{t}|\!|_{2}&\le r_u, ~1\!\le\! t\! \le\! K,\label{eq:control contraint}
\end{align}
\end{subequations}
where the optimization is over feasible controls, and equations \eqref{eq:riccati 1}-\eqref{eq:riccati 4} represent one iteration of the Riccati equation to calculate the first term of the objective.
\end{problem}

We now describe the performance of the above approaches. We perform several numerical simulations for various initial, process and observation uncertainties for both of the problems \ref{problem:Planning Problem OG} and \ref{problem:Planning Problem} and all three observation models. 

First, we provide an example for the range-squared observation model, where we show that the trajectory provided by the OG-based problem of \ref{problem:Planning Problem OG} can significantly under-perform in terms of reducing the estimation uncertainty. We show that planning based on the OG can result in undesirable trajectories for these partially observed problems, which stems from the fact that the OG is insensitive to the uncertainty parameters of the problem and provides the same result regardless of the changes in the three covariances. 

Next, we provide an example for the other model where qualitatively the output trajectories of the two problems resemble each other, but the covariance evolution results in the slight differences in the state trajectory contributing to a significant difference in the qualities of the trajectories in terms of the filters' performances. Lastly, we provide an example showing that when the intensity of noises tends to zero (particularly, if the sensor noise is very low), the performances of the OG-based and covariance-based trajectories tend to be close to each other. All our simulations are performed in MATLAB 2016b using the $ \mathtt{fmincon} $ solver.

For all the figures that depict the state trajectories:
\begin{itemize}
\item $\mathbf{x}\in\mathbb{R}^{2}$, $\mathbf{u}\in\mathbb{R}^{2}$, $\mathbf{z}\in\mathbb{R}$, and $ K=7 $;
\item $ \mathbf{W}^{u}_{t}=0\mathbf{I}_{2} $, $ r_u=0.8 $, $ r_g=0.1 $ and $ \mathbf{x}_g = (-1,2.25)^{T} $, which is indicated by a purple circle in the figures;
\item The units of the axes are in meters;
\item The initial estimate is $ \hat{\mathbf{x}}=(-1.5, -0.5)^{T} $, which is indicated by a green diamond in the figures;
\item The information sources are located at the centers of the light areas in the figures;
\item The initial trajectory for the solver, indicated with a dashed orange line, consists of three straight segments passing through $ (-1.5, -0.5)^{T} $, $ (-1.4, 0.21)^{T} $, $ (-1.1, 1.369)^{T} $, and $ (-1, 2.25)^{T} $. Hence, the deterministic system is observable for all three models; and
\item The optimized trajectory is shown by a solid cyan line.
\end{itemize}
\begin{figure}[t!]
\centering
   \subfloat[Range-squared, OG-Based\label{fig:rs OG cov 2}]{\includegraphics[width=0.5\linewidth]{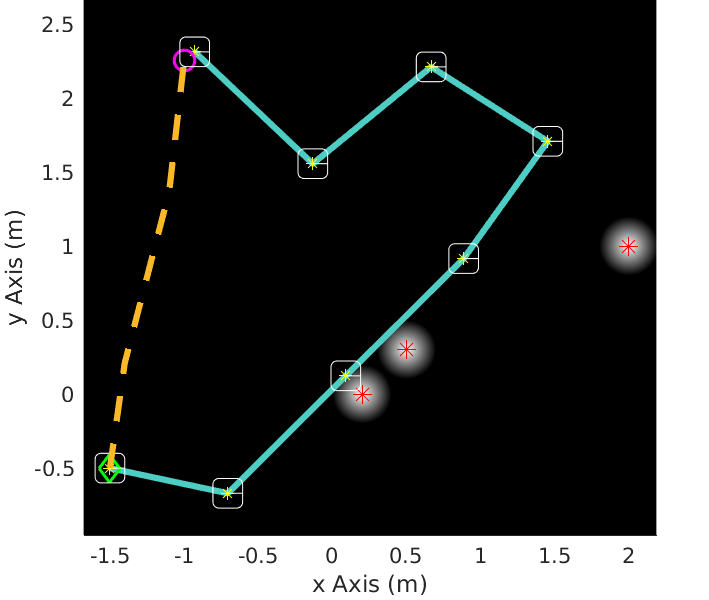}}
   \subfloat[Range-squared, Cov-Based\label{fig:rs filter cov 2}]{\includegraphics[width=0.5\linewidth]{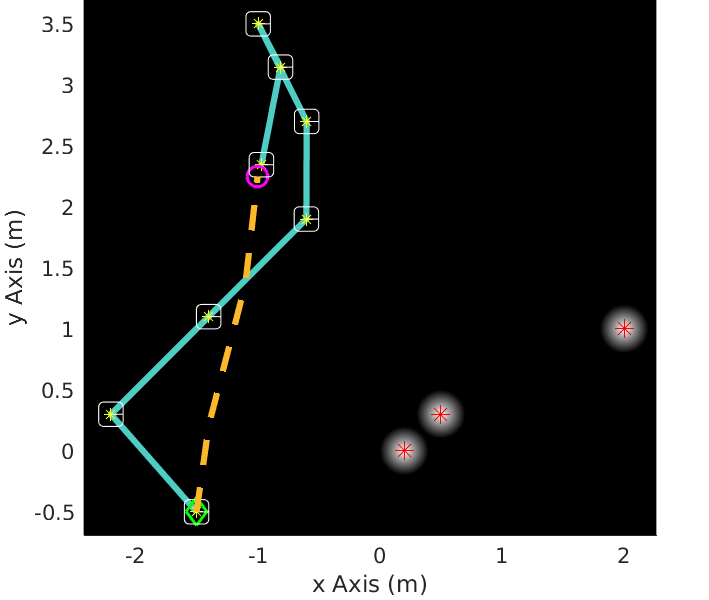}}
   \\\subfloat[Range, OG-Based\label{fig:r OG cov_3}]{\includegraphics[width=0.5\linewidth]{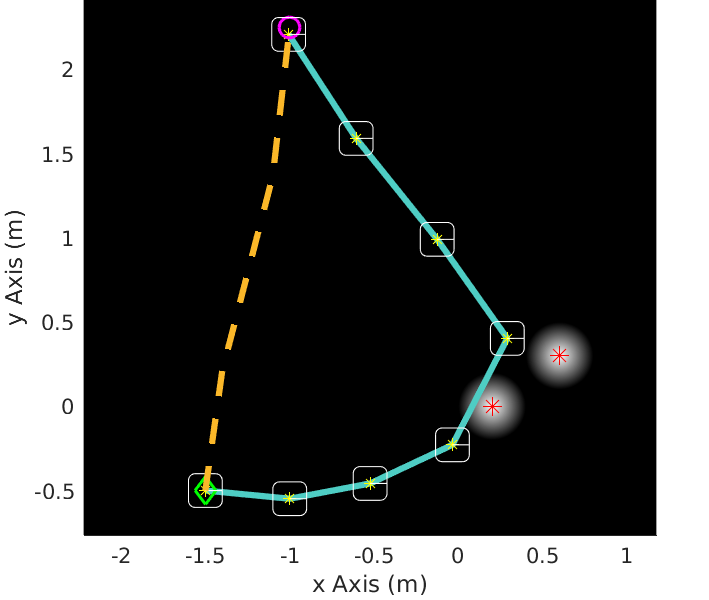}}
      \subfloat[Range, Cov-Based\label{fig:r filter cov 3}]{\includegraphics[width=0.5\linewidth]{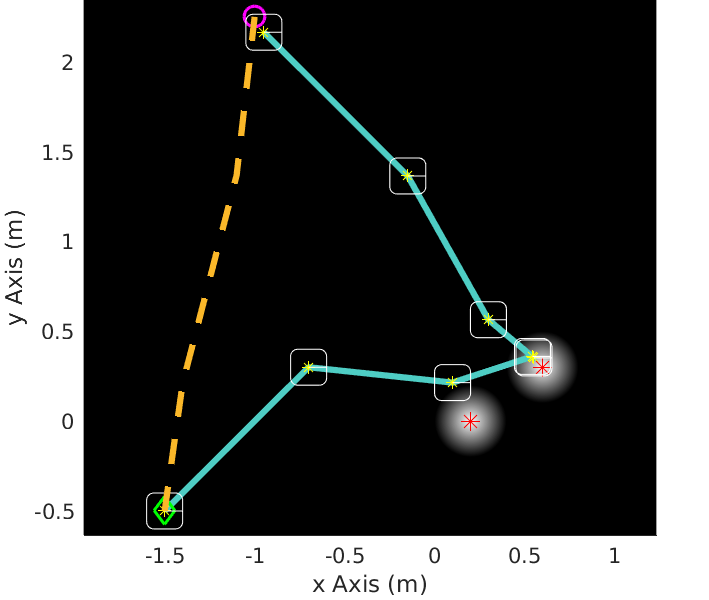}}
   \caption{Simulation results for the planning problem \ref{problem:Planning Problem OG} based on the condition number of the OG for range-squared and range observation models in (a) and (c), and the planning problem \ref{problem:Planning Problem} using  the trace of the covariance for range-squared and range observation models in (b) and (d), respectively. The information sources are located in the centers of the light areas. The dashed orange line represents the initial trajectory, while the solid cyan line shows the optimized trajectory.\label{fig:range-squared cov 2 figures}} 
\end{figure}
\begin{figure}[!]
\centering
   \subfloat[Range-squared\label{fig:rs cov 2}]{\includegraphics[width=0.5\linewidth]{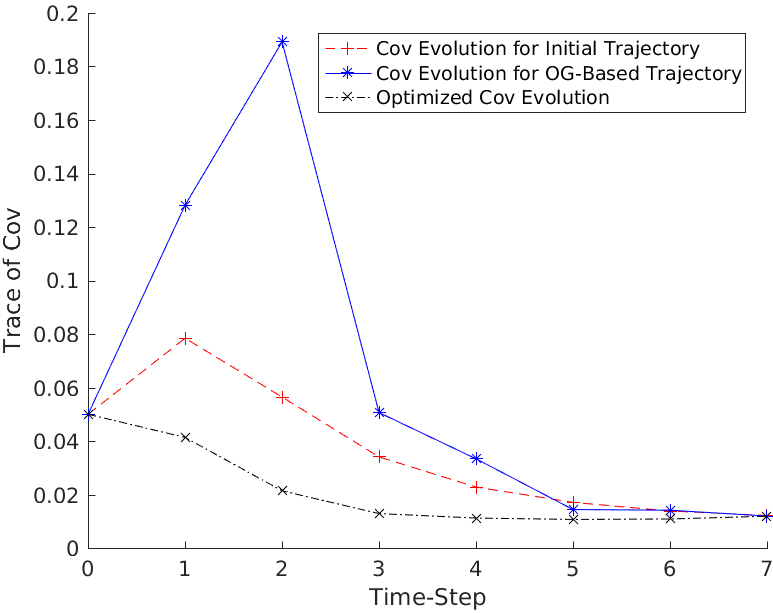}}
    \subfloat[Range\label{fig:r cov 3}]{\includegraphics[width=0.5\linewidth]{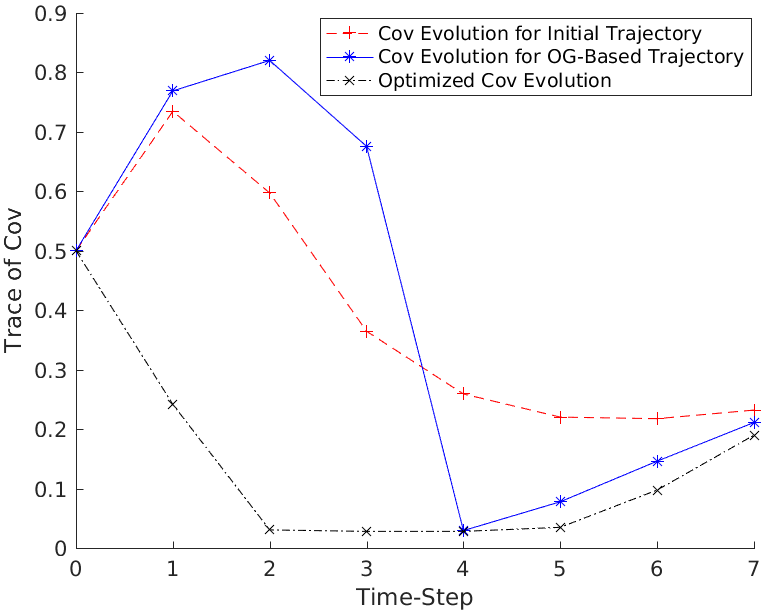}}
   \caption{Evolution of the trace of the covariance along the trajectory for the initial trajectory, optimization based on the OG measure, and optimization based on the covariance measure of the trajectories in Fig. \ref{fig:range-squared cov 2 figures}.}
\end{figure}

\subsection{Range-Squared-Only Observations}
Figures \ref{fig:rs OG cov 2} and \ref{fig:rs filter cov 2} show the results of the simulations for the range-squared-only observation model using the condition number of the OG and the trace of the covariance along the trajectory as the cost function, respectively. Information sources are at $ (0.2, 0)^{T} $, $ (0.5, 0.3)^{T} $, and $ (2, 1)^{T} $, and
\begin{align*}
\boldsymbol{\Sigma}_{\mathbf{x}_{0}}=\begin{pmatrix}
0.025 & 0.002\\ 0.002 & 0.025
\end{pmatrix}, 
\boldsymbol{\Sigma}_{\boldsymbol{\omega}}=\begin{pmatrix}
0.3 & 0.0\\ 0.0 & 0.1
\end{pmatrix},
\Sigma_{\nu}=0.1.
\end{align*}

Figure \ref{fig:rs cov 2} shows the evolution of the trace of covariance along the trajectories. While it is expected that the trajectory deigned based on the covariance evolution performs better than the other ones, it is surprising to observe that the OG-based trajectory actually under-performs the initial trajectory as well. Even though we have only shown the results of the simulation for the condition number of OG, the interested reader can find a more detailed set of experiments with other measures of the Gramian in a companion technical report \cite{r2017GramianExtended}, which parallel the results provided here. The quantitative result of Fig. \ref{fig:rs cov 2}, along with the qualitative difference in the trajectories as indicated in Fig. \ref{fig:range-squared cov 2 figures}, indicate that a measure of the OG is not a reliable measure to optimize in a problem with initial, process and observation uncertainties. 

\begin{figure}[t!]
\centering
   \subfloat[OG-Based Trajectory\label{fig:r OG cov_5}]{\includegraphics[width=0.5\linewidth]{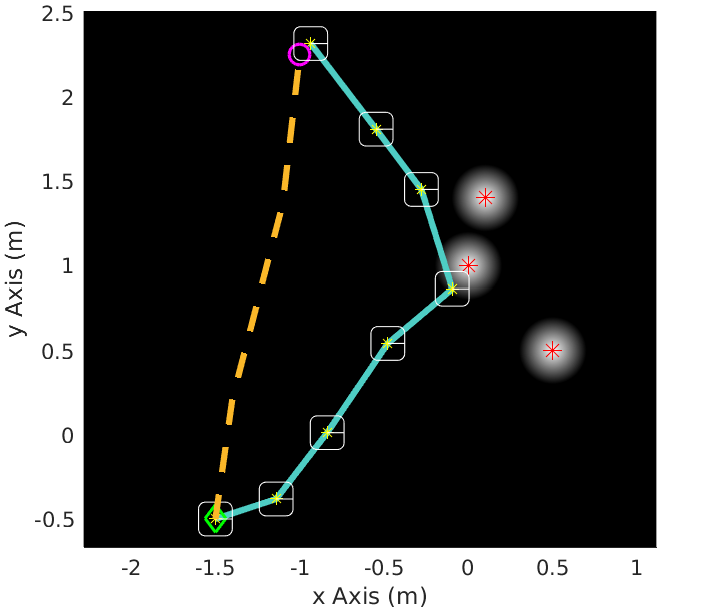}}
   \subfloat[Cov-Based Trajectory\label{fig:r filter cov 5}]{\includegraphics[width=0.5\linewidth]{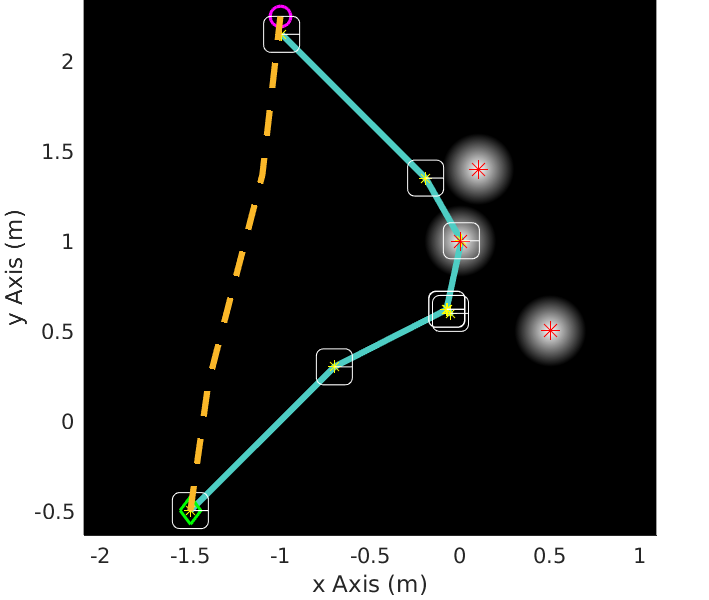}}
   \caption{Range-only observation model: a) The optimized state trajectory of the planning problem \ref{problem:Planning Problem OG} using the condition number of the OG as the cost function, b) The optimized state trajectory of the planning problem \ref{problem:Planning Problem} using the trace of the covariance as the cost function. The information sources are located in the centers of the light areas. The dashed orange line represents the initial trajectory, while the solid cyan line shows the optimized trajectory.\label{fig:range cov 5 figures}}
\end{figure}
\begin{figure}[!]
\centering
   \includegraphics[width=0.75\linewidth]{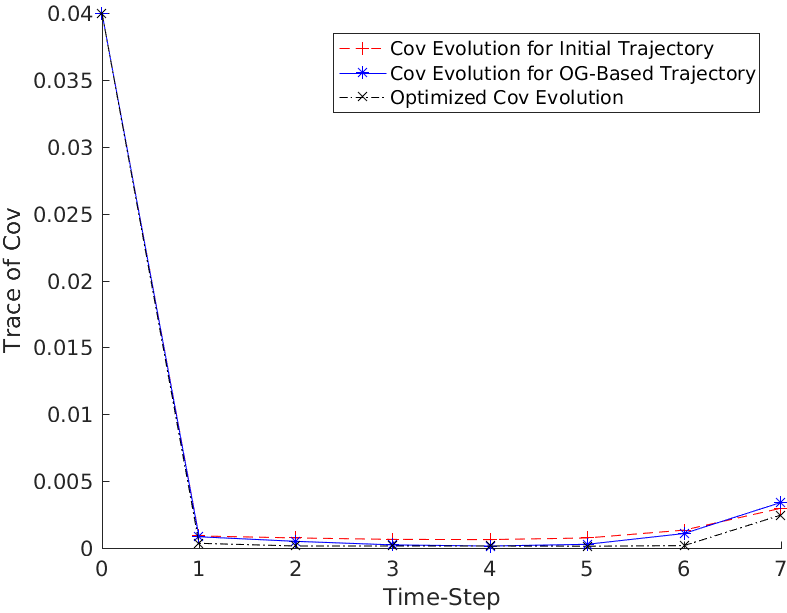}
   \caption{Range observation model. Evolution of the trace of the covariance along the trajectory for the initial trajectory, optimization based on the OG measure, and optimization based on the covariance measure of the trajectories in Fig. \ref{fig:range cov 5 figures}.\label{fig:r cov 5}}
\end{figure}
\subsection{Range-Only Observations}
Figures \ref{fig:r OG cov_3} and \ref{fig:r filter cov 3} show the results of the similar simulations for the range-only observation model with the condition number of the OG and the trace of the covariance as the cost function, respectively. Information sources are at $ (0.2, 0)^{T} $, and $ (0.6, 0.3)^{T} $, and
\begin{align*}
\boldsymbol{\Sigma}_{\mathbf{x}_{0}}=\begin{pmatrix}
0.25 & 0\\ 0 & 0.25
\end{pmatrix}, 
\boldsymbol{\Sigma}_{\boldsymbol{\omega}}=\begin{pmatrix}
0.1 & 0\\ 0 & 1
\end{pmatrix},
\Sigma_{\nu}=0.015.
\end{align*}
Figure \ref{fig:r cov 3} shows the covariance evolution for the trajectories of this simulation, which resembles the results of Fig. \ref{fig:rs cov 2}.

\subsection{Another Range-Only Scenario}
Last, Figs. \ref{fig:r OG cov_5} and \ref{fig:r filter cov 5} show the results of another set of simulations for the range-only observation model using condition number of the OG and the trace of the covariance, respectively. Information sources are located at $ (0, 1)^{T} $, $ (0.5, 0.5)^{T} $, and $ (0.1, 1.4)^{T} $, and
\begin{align*}
\boldsymbol{\Sigma}_{\mathbf{x}_{0}}=\begin{pmatrix}
0.02 & 0\\ 0 & 0.02
\end{pmatrix}, 
\boldsymbol{\Sigma}_{\boldsymbol{\omega}}=\begin{pmatrix}
0.1 & 0\\ 0 & 0.1
\end{pmatrix},
\Sigma_{\nu}=0.0001.
\end{align*}

In this experiment, the reduced noise covariances, particularly the observation covariance, lead to the high quality of measurements from a broad class of trajectories. As a result, the trace of covariance evolution of Fig. \ref{fig:range cov 5 figures} indicates only a slight difference between the three trajectories.

\textit{Remark:} It should be noted that in all the figures, since the state trajectories are softly constrained to reach to the same goal region at the end of the navigation, the covariance evolutions converge to each other towards the end of the trajectories. This is due to the fact that in the Bayesian filtering, the latest observations (which arise from the same region in the state space) carry a higher weight than the prior history. As a result, in comparing the covariance evolutions, the variations in the behavior along the entire trajectory is of concern since a highly certain trajectory can lead to safer navigation, particularly, in a complex environment with obstacles, banned areas or multiple agents.

\textit{Remark:} Finally, note that the simulation times to solve the optimization problem for all cases are of the same order, which stems from the fact that the computation complexity of both the problems \ref{problem:Planning Problem OG} and \ref{problem:Planning Problem} is $ O(Kn_x^3) $ \cite{rafieisakhaei2016belief}.

\section{Conclusion}
In this paper, we have investigated a well-known heuristic employing the observability Gramian in planning under observation uncertainty. We have utilized two common observation models and shown that, in general, the observability Gramian (and the closely-related standard Fisher information matrix) fail to capture many aspects of the models including the initial, process, and observation uncertainties. As a result, based on changes in those models, we showed using analytic and numerical examples that planning based on the observability Gramian can provide trajectories that are very different in terms of the estimation performance from the optimal plans based on the estimation covariance of the problem.

\bibliographystyle{IEEEtran}
\bibliography{MohammadRaf.bib}
\end{document}